\pdfoutput=1
\documentclass[10pt,journal]{IEEEtran}
\usepackage{amsmath,amssymb}
\usepackage{graphicx}
\usepackage{booktabs}
\usepackage{cite}
\usepackage[hidelinks]{hyperref}

\title{ReSUL: Learning Regional Sensor Utility for Robust
Multisensor Semantic Segmentation}
\author{Yiwen~Wang and Jiahao~Qin%
\thanks{Corresponding author: Jiahao Qin (jiahao.qin19@gmail.com).}}
\hypersetup{pdftitle={ReSUL: Learning Regional Sensor Utility for Robust Multisensor Semantic Segmentation},pdfauthor={Yiwen Wang and Jiahao Qin}}
\begin{document}
\maketitle
\begin{abstract}
A corrupted sensor observation is not necessarily useless: misalignment
can place informative scene structure at the wrong location. Robust
multisensor segmentation must determine where to retain this evidence,
where to exclude it, and where repair can recover its value.
We introduce Regional Sensor Utility Learning (ReSUL), which learns these
choices through their effect on segmentation risk. ReSUL compares full-sensor
predictions, sensor-exclusion paths, and a re-registration path for
displaced thermal inputs. A router predicts each path's signed reduction
in category-balanced Brier risk relative to a fixed reference, then uses
these estimates to allocate spatial mixture weights. This supervision
connects spatial allocation to measurable task benefit without requiring
fault labels at inference. MFNet experiments compare project-trained
models under eleven clean and disturbed input conditions.
ReSUL ranks first in 10 of 11 conditions in both ImageNet-1k and
ImageNet-22k comparisons. Under severe thermal displacement, it reaches
54.58\% and 55.63\% mIoU, exceeding the strongest competing method in
each group by 7.19 and 5.63 percentage points.
Fixed-bank MFNet ablations show that utility supervision improves all
eleven condition means over task-only routing. These results support
learning how to use disturbed observations through their regional
prediction benefit.
Code is available at \url{https://github.com/D-ST-Sword/ReSUL}.
\end{abstract}
\begin{figure*}[t]
\centering
\includegraphics[width=\textwidth]{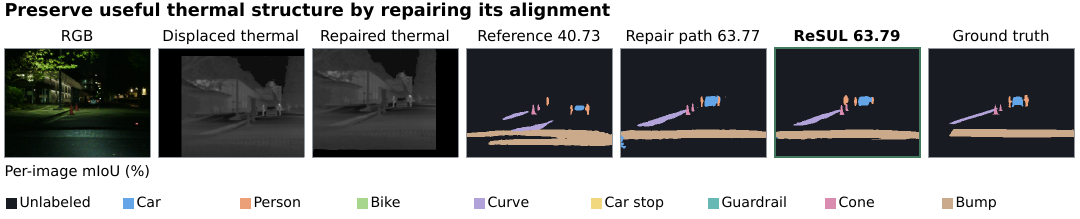}
\caption[Regional correction of thermal misalignment]{MFNet under severe thermal displacement, using the project-trained R1 bank. Observed and re-registered thermal inputs, reference and repair-path predictions, ReSUL, and ground truth retain native aspect ratios. Numbers denote per-image mIoU (\%) over nonzero-union classes, including class 0; colors are shared with Fig.~\ref{fig:qualitative}. Re-registration restores useful thermal evidence.}
\label{fig:teaser}
\end{figure*}
\section{Introduction}
\label{sec:introduction}

Multisensor segmentation draws on complementary observations to parse
scenes when visible images alone are ambiguous~\cite{mfnet,muses}.
The value of these observations depends on their condition, making
sensor quality a central concern for deployed multimodal
systems~\cite{liu2025deployment,zhang2026lowquality}. Complementary inputs
can also introduce semantic conflicts~\cite{zhao2025scrnet}. A displaced
thermal boundary may support the right category at the wrong location. Discarding
the observation avoids this conflict but also loses recoverable scene
information; retaining it without correction can propagate the error.
The question is therefore how to use a disturbed observation, not simply
whether it is present or corrupted.

Fusion networks learn cross-modal interactions, spatial balancing, and
condition-dependent combinations~\cite{cmx,spatialseg,peafusion,cafuser,dgfusion}.
Adaptive selection varies modality contributions by category and
location~\cite{adapnet,ssma}, while reliability-aware fusion uses
uncertainty or historical losses~\cite{uno,qmf}. Masked training improves
tolerance to missing inputs~\cite{crm}. Our focus is the regional task
benefit of different responses to the same observation. A confidence
score alone does not specify whether excluding that observation or
repairing its geometry would yield a better segmentation. Comparing
these alternatives makes the choice directly learnable from their
prediction errors.

We propose Regional Sensor Utility Learning (ReSUL) to learn where
preserving, excluding, or repairing an observation improves segmentation
(Fig.~\ref{fig:teaser}). These choices are represented by prediction paths.
Full-sensor paths retain
cross-modal information; sensor-exclusion paths remove an observation;
and the MFNet re-registration path restores displaced thermal structure.
A regional router learns each path's reduction in category-balanced
Brier risk relative to a fixed full-sensor reference. Positive and
negative utilities encode improvement and deterioration, giving the
router a direct learning signal for mixing predictions. The bank provides
ways to respond to interference, while utility learning determines where
they should contribute. Synthetic faults supply training examples;
inference uses sensor observations and predictions without condition labels.

We make three contributions. First, we formulate robust fusion as learning
the regional task utility of retaining, excluding, or repairing sensor
observations. Full-sensor, restricted-input, and repaired-input prediction
paths make these choices explicit, and spatial allocation preserves useful
evidence where blanket sensor rejection would discard it.

Second, we learn allocation through signed regional risk reduction.
The target distinguishes beneficial from harmful alternatives, and
full and restricted paths from the same network quantify the task
benefit of sensor exclusion. This links routing to prediction quality
rather than leaving the allocation scores without direct supervision.

Third, MFNet experiments demonstrate robustness to geometric and
appearance disturbances with project-trained models. ReSUL ranks first
in 10 of 11 conditions in each initialization group. Matched-bank
ablations, exclusion calibration, and regional visualizations establish
how the available paths and utility learning contribute to these results.
\section{Related Work}
\label{sec:related_work}

\subsection{Multisensor Semantic Segmentation}
RGB--thermal networks retain modality-specific structure while exchanging
complementary features. RTFNet~\cite{rtfnet} uses dual-stream encoding,
ABMDRNet~\cite{abmdrnet} reduces modality differences, and CMX~\cite{cmx}
combines feature rectification with cross-modal fusion. CMNeXt~\cite{deliver}
extends information selection to arbitrary sensor combinations.
Recent approaches include PEAFusion's frozen-backbone
adapters~\cite{peafusion}, SpatialSeg's spatial balancing~\cite{spatialseg},
Sigma's Siamese state-space encoders~\cite{sigma2025}, and AMDANet's
attention-driven discrepancy alignment and semantic-consistency
inference~\cite{amdanet2025}.

Several recent methods address how complementary information enters the
segmentation network. SCRNet~\cite{zhao2025scrnet} rectifies semantic
conflicts, while RXSSNet~\cite{zhang2026rxssnet} reweights features before
local--global aggregation. MCLNet~\cite{lu2026mclnet} jointly learns image
fusion and segmentation, and FGDNet~\cite{zhao2025fgdnet} transfers
multimodal and complementary features through generative distillation.
IQSeg~\cite{liu2026iqseg} couples implicit deformable alignment with query
refinement to handle RGB--thermal misalignment. Cross-domain registration
also separates appearance variation from geometry: PCReg-Net~\cite{qin2026pcreg}
uses progressive contrast-guided refinement in retinal and microscopy
images. ReSUL makes a repaired prediction one of several selectable
responses, so registration quality is judged through its regional
segmentation benefit.

For multisensor driving perception, CAFuser~\cite{cafuser} conditions
fusion on the environment, DGFusion~\cite{dgfusion} uses depth guidance,
and MM SAM-Adapter~\cite{mmsam2025} selectively incorporates auxiliary
features into pretrained RGB representations. In RGB--thermal image
fusion, SGAFuse~\cite{yang2026sgafuse} combines semantic attention with
dynamic multi-path gates. These architectures adapt the exchange or
combination of features. ReSUL allocates final predictions from full,
restricted, and repaired inputs using an explicit task-risk target.

\subsection{Input Reliability and Robust Fusion}
Low-quality multimodal learning encompasses noisy, incomplete, imbalanced,
and changing observations~\cite{zhang2026lowquality}.
DeLiVER~\cite{deliver}, MUSES~\cite{muses}, and the sensor-failure benchmark
of Liao et al.~\cite{sensorfailurebenchmark} support evaluation across
sensor combinations and degraded observations. The iMML
framework~\cite{lopez2026incomplete} further distinguishes missingness
patterns when evaluating incomplete multimodal learning.
Complementary random masking~\cite{crm} reduces modality dependence
through masked training and self-distillation. RAML~\cite{chen2025raml}
uses cross-modal redundancy and element-wise feature-quality estimates
to retain uncorrupted information. Reconstruction offers another route:
DCER~\cite{wang2026dcer} combines frequency and bottleneck compression
with energy-based missing-feature recovery, while
ProgDiff~\cite{dai2026progressive} uses conditional diffusion to recover
distribution-consistent modality representations. ReSUL's repair path
instead corrects the geometry of an observed thermal image and competes
with retention and exclusion paths.

Reliability-aware fusion estimates which inputs deserve influence.
UNO~\cite{uno} scales predictions through uncertainty and spatial
temperature estimates before noisy-or fusion, and QMF~\cite{qmf}
regularizes fusion-weight ordering using historical losses.
DUAL~\cite{qin2025dual} models feature uncertainty and uncertain
cross-modal relationships. HTEF~\cite{yang2025htef} aggregates
hierarchical evidence with opinion-consistency learning;
ECoLaF~\cite{ecolaf2025} discounts conflicting evidence, and
ReCoLaF~\cite{recolaf2026} combines class-wise reliability with pixel-level
conflict adjustment. For driving perception, RAF~\cite{raf2026} supervises
camera reliability through cross-modal similarity and cleanliness
annotations, whereas KAME~\cite{shi2026kame} supplies sensor-reliability
rules as language-level context for frame--event reasoning.

Detecting unfamiliar inputs and calibrating confidence are related but
distinct objectives. MOAL~\cite{wang2025moal} learns out-of-distribution
awareness from multi-view representations. Cheon and
Paik~\cite{cheon2026calibration} improve confidence calibration through
random-noise warm-up. ReSUL supervises a different quantity: the signed
change in regional risk when a specific alternative replaces the
reference. This target distinguishes whether exclusion or repair helps
the task, beyond whether an input appears unusual or a prediction is
confident.

\subsection{Adaptive Experts and Task-Relevant Information}
The organization of multimodal representations determines which
information survives fusion. BC-PMJRS~\cite{qin2025bcpmjrs} balances
cross-modal redundancy and task discrimination in a joint representation,
while MoIF~\cite{qin2026moif} regulates interaction orders through
modality-subset predictions in humor detection. In cellular
representation learning, Richter et al.~\cite{richter2026synergy}
separate tasks that benefit from cross-modal synergy from those already
served by a strong unimodal representation. These perspectives motivate
assessing fusion by the information it contributes to a particular
prediction, rather than by interaction strength alone.

Input-dependent assignment is central to mixtures of experts~\cite{moe}.
Hierarchical adaptive experts~\cite{qin2025hae} combine global and local
modality representations for sentiment analysis.
AdapNet~\cite{adapnet} and self-supervised model adaptation~\cite{ssma}
apply adaptive modality selection across locations and categories;
CMFS~\cite{cmfs2025} combines CLIP-guided regional interaction with a
low-rank mixture of experts. Learning to defer compares a predictor
with an alternative decision maker~\cite{defer}; multiple-expert
extensions study correctness estimation, calibration, and ensemble
selection~\cite{multil2d}. Our dense prediction setting supervises
continuous, category-balanced regional risk, giving allocation scores
an explicit meaning as gains or losses against a reference prediction.
\section{Regional Sensor Utility Learning}
\label{sec:method}

\begin{figure*}[t]
\centering
\includegraphics[width=\textwidth]{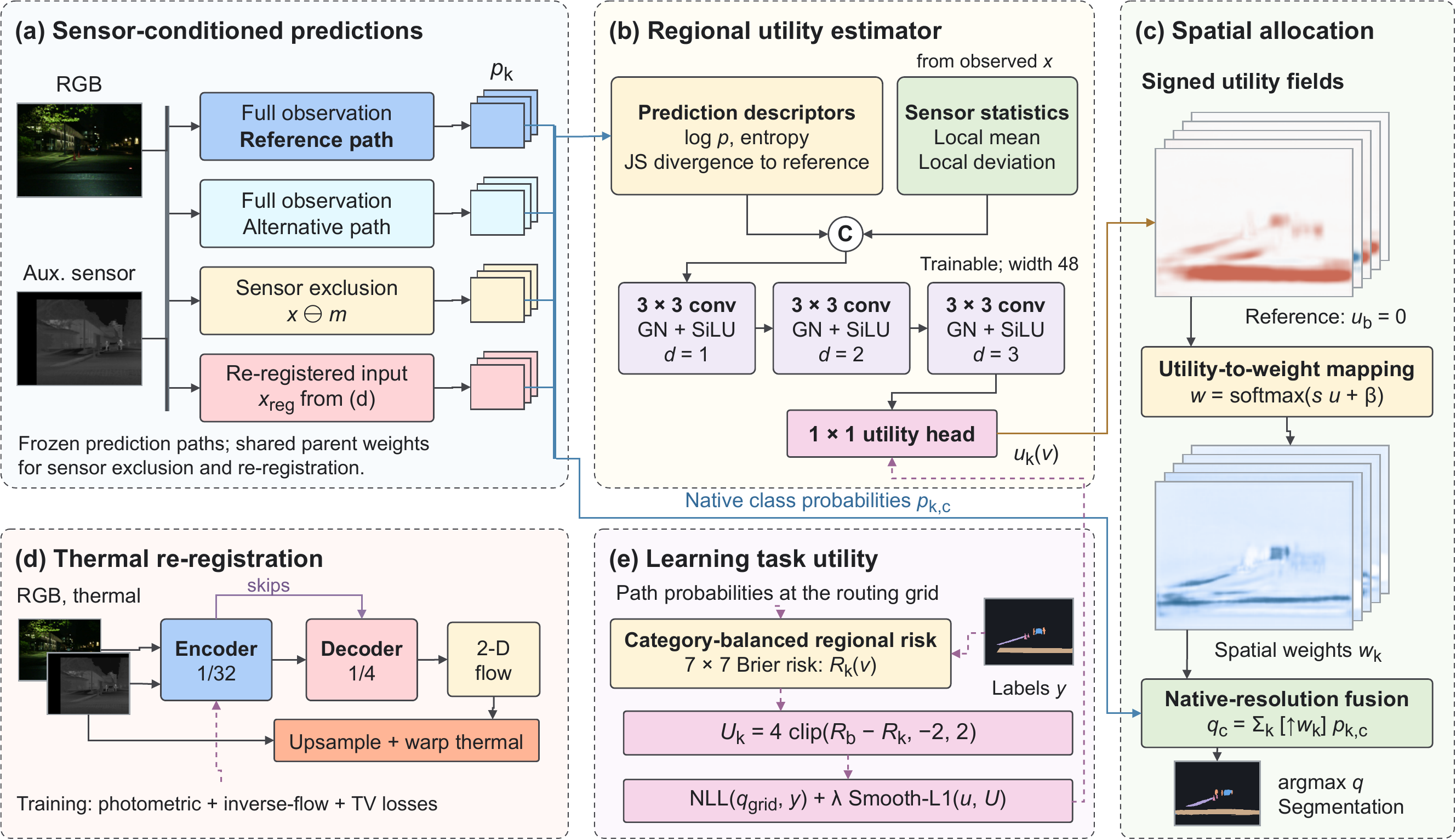}
\caption[ReSUL architecture]{ReSUL architecture. Sensor-conditioned prediction paths (a) feed a regional utility estimator (b) and native-resolution probability fusion (c). Insets show thermal re-registration (d) and risk-based supervision (e); dashed arrows denote training-only signals. Tiles use the project-trained MFNet R1 example: red/blue denote positive/negative utility, and darker blue denotes larger weight. Only spatial weights are upsampled for fusion.}
\label{fig:architecture}
\end{figure*}

ReSUL learns the task value of sensor observations through the prediction
paths that retain, exclude, or repair them (Fig.~\ref{fig:architecture}).
Regional risk reduction relative to a fixed reference defines a common
utility for these alternatives. A utility estimator predicts this quantity
from the observed inputs and path predictions, then converts it into
spatial mixture weights. The prediction-bank configurations are specified
in Sec.~\ref{sec:settings}.

\subsection{Prediction Alternatives}
Let $x=(x^{(1)},\ldots,x^{(M)})$ denote the sensor observations and
$p_k(v,c)$ the probability that path $k$ assigns to category $c$ at
routing-grid location $v$. All $K$ paths predict the same categories in
the same image coordinates, and one path, indexed by $b$, is the
\emph{reference path} for every input condition. Multiple fusion networks,
with potentially different architectures, can contribute to the bank.

For a network $F_\theta$, its full-sensor path is
$p_{\mathrm{full}}^\theta=F_\theta(x)$ and its sensor-exclusion path is
$p_{\setminus m}^\theta=F_\theta(x\ominus m)$, where $x\ominus m$
zeros sensor $m$ before normalization. These paths share parameters but
produce explicit alternatives for retaining or excluding an observation.
On MFNet, $p_{\mathrm{reg}}^\theta=F_\theta(x_{\mathrm{reg}})$ adds a
third option: $x_{\mathrm{reg}}$ replaces the thermal input with its
re-registered version (Sec.~\ref{sec:realign}). A full-sensor path is not
necessarily the reference; the bank configuration fixes $b$.
Probabilities are sampled onto a common routing grid, while the final
mixture uses native-resolution maps.

\subsection{Supervising Regional Utility}
Utility supervision asks whether an alternative improves the prediction
in a particular region. We compare paths using category-balanced regional
Brier risk~\cite{brier1950}, so the target accounts for both their errors
and the affected categories. Let
$y_c(z)$ be the one-hot annotation, $m(z)$ indicate a valid label, and
$\mathcal N(v)$ denote a $7\times7$ neighborhood on the routing grid.
For $M(v)=\sum_{z\in\mathcal N(v)}m(z)>0$, define
\begin{equation}
 R_k(v)=\frac{1}{M(v)}\sum_{z\in\mathcal N(v)}m(z)
             \sum_{c=1}^{C}a_c[p_k(z,c)-y_c(z)]^2.
 \label{eq:risk}
\end{equation}
For clean fitting-label frequencies $f_c$, we set
$t_c=\min\{20,\max(f_c,10^{-4})^{-1/2}\}$ and
$a_c=t_c/(C^{-1}\sum_jt_j)$, giving mean-one weights. Balancing reduces the
influence of frequent categories, and the one-versus-rest terms penalize
false foreground predictions even in otherwise correct neighborhoods.

The utility target compares each path with the reference:
\begin{equation}
 U_k(v)=4\,\operatorname{clip}\bigl(R_b(v)-R_k(v),-2,2\bigr).
 \label{eq:utility-target}
\end{equation}
Positive utility denotes lower standalone regional risk. Neighborhood
averaging supplies context around boundaries and small objects; empty
neighborhoods are excluded. Targets use detached path predictions and
fitting labels and remain fixed during router optimization.

Same-parent paths define the exclusion benefit
$H_m^\theta(v)=R_{\mathrm{full}}^\theta(v)
-R_{\setminus m}^\theta(v)$. Positive values mean that removing sensor
$m$ lowers that network's risk; negative values mean that retaining it
helps. Before clipping, the difference of their reference-relative
targets equals $4H_m^\theta$, even when their parent is not the reference.
The prediction $(u_{\setminus m}^\theta-u_{\mathrm{full}}^\theta)/4$
estimates this contrast; target clipping can limit its magnitude.

\subsection{Regional Utility Router}
The router estimates utility from prediction disagreement and local
sensor statistics, allowing the allocation to vary within a scene.
Path probabilities are floored at $10^{-6}$ and renormalized across
categories. The router concatenates their log probabilities (clipped to
$[-12,0]$ and divided by six), their entropy,
their Jensen--Shannon divergence from the reference, and local
sensor-channel means and standard deviations. Neither semantic labels nor
disturbance identities are inputs. Three width-48 $3\times3$ convolutional
layers with dilations one, two, and three (eight-group normalization, SiLU) encode
the descriptors, and a $1\times1$ head predicts one utility field
$u_k(v)$ per path, shared across categories, with the reference fixed at
$u_b(v)=0$. The utilities determine a distribution over paths,
\begin{equation}
 w_k(v)=\frac{\exp(su_k(v)+\beta_k)}
                  {\sum_j\exp(su_j(v)+\beta_j)},
 \label{eq:weights}
\end{equation}
with a fixed routing scale $s$ and fixed priors $\beta_k$, and the
mixture probability is
\begin{equation}
 q_c(v)=\sum_k w_k(v)p_k(v,c).
 \label{eq:mixture}
\end{equation}
At native resolution the weight fields are bilinearly interpolated and
applied to the original path probabilities; the final category is the
argmax of this mixture.

The router learns both to produce an accurate mixture and to predict
each alternative's task benefit. Training combines the segmentation
objective with direct utility regression,
\begin{equation}
 \mathcal L=\mathcal L_{\mathrm{seg}}+
             \lambda\mathcal L_{\mathrm{utility}},
 \label{eq:objective}
\end{equation}
where $\mathcal L_{\mathrm{seg}}$ is category-weighted negative log
likelihood over valid routing-grid labels and
$\mathcal L_{\mathrm{utility}}$ averages the Smooth-L1 error between
$u_k$ and $U_k$ over non-reference paths and nonempty neighborhoods.
Bank parameters remain frozen during router fitting. Development data
determine the checkpoint and fixed hyperparameters.

\paragraph{From path utility to mixture risk}
The quadratic risk clarifies why the two objectives are complementary.
At a location, write $\ell(p,y)=\|p-y\|_a^2$, where
$\|r\|_a^2=\sum_c a_c r_c^2$. Expanding the squares for
$q=\sum_k w_kp_k$ gives
\begin{equation}
\begin{aligned}
 \ell(p_b,y)-\ell(q,y)
 &=\sum_k w_k[\ell(p_b,y)-\ell(p_k,y)]\\
 &\quad+\sum_k w_k\|p_k-q\|_a^2.
\end{aligned}
\label{eq:mixture-risk}
\end{equation}
The first term weights individual path benefits; the second accounts
for their combination through weighted prediction disagreement.
The identity holds pointwise with spatially varying weights and remains
exact when both sides are averaged over valid neighborhoods.
Our smoothed, clipped utility targets teach the local value of each
alternative, while the segmentation loss directly optimizes their
combined prediction. Thus utility regression supplies path-level
supervision alongside the mixture-level training signal.

\subsection{Repairing Displaced Thermal Evidence}
\label{sec:realign}
Exclusion removes an observation from a prediction, whereas
re-registration makes displaced scene information available for fusion.
On MFNet, AlignNet predicts a displacement field from RGB and the observed
thermal image. A width-32 encoder--decoder encodes to $1/32$ resolution
and decodes to $1/4$ resolution. Its two-channel head applies
$0.5\tanh(\cdot)$ in normalized grid coordinates; bilinear upsampling
and grid sampling then warp the thermal image with zero padding.
The resulting prediction is one bank alternative, with its contribution
determined by the same utility router.

Training starts from clean MFNet pairs and synthesizes identity, global
shift, rotation, or quarter-image shift with probabilities
$0.25,0.35,0.20,0.20$. Known inverse fields and recoverable-pixel masks
supervise reconstruction. The loss is
$\mathcal L_{\mathrm{align}}=\mathcal L_{\mathrm{photo}}
+2\mathcal L_{\mathrm{flow}}+0.1\mathcal L_{\mathrm{TV}}$:
the first two terms are masked L1 errors for reconstructed thermal
intensity and the inverse field, respectively, and the last is the sum
of mean absolute horizontal and vertical field differences. Synthetic
geometry supplies the training target; inference estimates the field
from the sensor observations alone.
\section{Experiments}
\label{sec:experiments}
\subsection{Datasets and Evaluation}
MFNet~\cite{mfnet} provides RGB--thermal images and nine-category
annotations. We fit, select, and evaluate routers on its official
784/392/393 training/validation/test split. Native scoring uses the
$480\times640$ labels, including the unlabeled category in the
nine-category mean.

We report native-resolution mean intersection over union (mIoU) and
class IoUs. MFNet results use mean $\pm$ sample SD over three/five
runs for ImageNet-1k/22k initialization. Each ReSUL run averages three
router initializations; aggregation details are given in Appendix~\ref{app:training}.

\subsection{Sensor Disturbances}
Disturbances act after geometric mapping and before checkpoint-specific
normalization. MFNet inputs are $480\times640$. Methods share disturbance
parameters and original annotations.

MFNet router fitting uses clean inputs, RGB darkening, RGB noise, and
thermal displacement.
Darkening scales camera intensities by 0.15. Moderate displacement
shifts an auxiliary observation by 8\%
of its width and 4\% of its height, with deterministic image-specific
horizontal direction and zero-filled boundaries; severe displacement
doubles the offsets. Camera noise has standard deviation 0.25 in MFNet's
$[0,1]$ range, with clipping.
Evaluation also includes camera blackout and MFNet thermal blackout.

Additional evaluation conditions include Gaussian camera blur (standard
deviation 0.008 times the shorter input dimension), $\pm10^\circ$ auxiliary
rotation, and localized auxiliary displacement or camera blackout in a
deterministic quarter-image rectangle. Outside pixels are unchanged.
These additional transformations are held out from router fitting;
rotation and localized displacement are included in AlignNet and
fault-augmented expert training. Condition labels and corruption masks
are not inputs to the router at inference.

\subsection{Prediction Banks and Training}
\label{sec:settings}
MFNet routers train for 30 epochs with frozen experts, batch
size 12, AdamW~\cite{adamw}, learning rate $3\times10^{-4}$, weight decay
$10^{-4}$, cosine decay to $3\times10^{-5}$, and random horizontal
flips. Routing uses a $120\times160$ grid,
$7\times7$ targets, zero priors, $s=10$, and $\lambda=5$.
Task-only training sets $\lambda=0$. Checkpoints maximize
$\overline{\mathrm{IoU}}_{\mathrm{dev}}
-2\max\{0,\mathrm{IoU}_{b,\mathrm{clean}}-0.001
-\mathrm{IoU}_{q,\mathrm{clean}}\}$, averaging the fitting conditions
on the routing grid in fractional units. Category frequencies use clean
routing-grid labels.

MFNet trains AlignNet for 60 epochs with batch
size 6, AdamW, learning rate $2\times10^{-4}$, weight decay $10^{-4}$,
and cosine decay. Its validation criterion minimizes mean nonidentity
endpoint error plus twice the identity endpoint error, both in pixels.

Two SpatialSeg-architecture networks~\cite{spatialseg}
use ImageNet-1k or ImageNet-22k initialization and the official MFNet
training split. The clean-trained $F_{\theta_c}$ supplies the reference
$p_b=F_{\theta_c}(x)$. The fault-augmented $F_{\theta_a}$ supplies four
alternatives: full inputs, no thermal, no RGB, and re-registered thermal.
Exclusion contrasts therefore pair paths within $F_{\theta_a}$.
Each expert receives 165,000 sample presentations. The reference uses
standard augmentation; $F_{\theta_a}$ additionally samples thermal
displacement, rotation, localized displacement, and blackout, along with
camera darkening, noise, and blackout.

CMX-B2, CMNeXt-B4, AMDANet, and PEAFusion-B/L are trained with their official
implementations and method-specific training configurations. Comparisons
in Table~\ref{tab:retrained} share the official data splits,
validation-based checkpoint selection, evaluation disturbances, and native
scoring, and are grouped by ImageNet initialization. Detailed training
configurations are given in Appendix~\ref{app:training}.

\paragraph{Controls and calibration} MFNet controls use equal-weight
averaging, remove utility supervision ($\lambda=0$), or remove
re-registration or sensor-exclusion paths and refit the router.
The remaining fitting and selection settings are unchanged.
Table~\ref{tab:factors} compares these variants; Fig.~\ref{fig:transitions}
reports every condition in the first ImageNet-1k run (R1).
Table~\ref{tab:supervision} compares entropy, task-only, and utility
signals over seven MFNet core conditions, pairing the fault-augmented
network's full and restricted paths. Calibration metrics use valid
neighborhoods with $|H_m|\geq0.0025$; RMSE uses raw Brier units.
The entropy baseline uses the neighborhood-averaged full-minus-restricted
entropy, with positive values favoring exclusion.
Positive and negative recall measure recognition of beneficial exclusion
and retention, respectively. Counts are pooled before computing each
run's metrics. Appendix~\ref{app:utility} includes the MFNet
eleven-condition aggregates.
\subsection{Clean Segmentation}
\label{sec:results}
\begin{table}[t]
\centering
\caption{Clean MFNet test results. All task models are trained in this project. Values are mean $\pm$ SD over 3/5 runs for ImageNet-1k/22k. Best and second-best means within each group are bold and underlined.}
\label{tab:clean}
\footnotesize
\setlength{\tabcolsep}{2pt}
\begin{tabular*}{\columnwidth}{@{\extracolsep{\fill}}llc@{}}
\toprule
Method & Backbone & mIoU (\%) \\
\midrule
\multicolumn{3}{l}{ImageNet-1k} \\
CMX-B2~\cite{cmx} & MiT-B2 & $\mathbf{58.17}\!\pm\!0.36$ \\
CMNeXt-B4~\cite{deliver} & MiT-B4 & $56.91\!\pm\!0.54$ \\
AMDANet~\cite{amdanet2025} & MiT-B4 & $57.08\!\pm\!1.45$ \\
PEAFusion-B~\cite{peafusion} & Swin-V2-B & $56.44\!\pm\!1.72$ \\
SpatialSeg (clean)~\cite{spatialseg} & Swin-B & $\underline{57.44}\!\pm\!0.36$ \\
ReSUL & Swin-B & $57.38\!\pm\!0.52$ \\
\midrule
\multicolumn{3}{l}{ImageNet-22k} \\
PEAFusion-B~\cite{peafusion} & Swin-V2-B & $56.63\!\pm\!1.65$ \\
PEAFusion-L~\cite{peafusion} & Swin-V2-L & $\underline{58.19}\!\pm\!1.67$ \\
SpatialSeg (clean)~\cite{spatialseg} & Swin-B & $57.56\!\pm\!1.55$ \\
ReSUL & Swin-B & $\mathbf{58.31}\!\pm\!0.51$ \\
\bottomrule
\end{tabular*}
\end{table}
ReSUL retains the clean-input accuracy of its full-sensor reference
while adding alternatives for disturbed observations
(Table~\ref{tab:clean}). With ImageNet-1k initialization, its 57.38\%
mIoU is within 0.06 percentage point of the clean-trained SpatialSeg
reference; CMX-B2 leads this group at 58.17\%. With ImageNet-22k
initialization, ReSUL achieves the highest mean, 58.31\%, compared with
58.19\% for PEAFusion-L and 57.56\% for the reference. The robustness
comparison below tests how these models use the same observations
after sensor disturbance.

\subsection{Robustness to Sensor Disturbances}
\begin{table*}[t]
\centering
\caption{MFNet robustness with project-trained models (mIoU, \%). All methods use mean $\pm$ SD over 3/5 runs for ImageNet-1k/22k. SpatialSeg (clean) is the full-sensor reference. Rankings follow Table~\ref{tab:clean}.}
\label{tab:retrained}
\footnotesize
\setlength{\tabcolsep}{2pt}
\begin{tabular*}{\textwidth}{@{\extracolsep{\fill}}l*{6}{c}@{}}
\toprule
\multicolumn{7}{l}{\textit{Original inputs and camera perturbations}} \\
Method & Clean & Dark & Noise & Off & Blur & Local off \\
\midrule
\multicolumn{7}{l}{ImageNet-1k} \\
CMX-B2~\cite{cmx} & $\mathbf{58.17}\!\pm\!0.36$ & $54.02\!\pm\!0.96$ & $51.47\!\pm\!0.44$ & $46.13\!\pm\!0.75$ & $52.07\!\pm\!0.39$ & $53.05\!\pm\!0.20$ \\
CMNeXt-B4~\cite{deliver} & $56.91\!\pm\!0.54$ & $55.26\!\pm\!0.28$ & $44.27\!\pm\!0.17$ & $33.61\!\pm\!2.38$ & $48.37\!\pm\!0.81$ & $48.51\!\pm\!0.89$ \\
AMDANet~\cite{amdanet2025} & $57.08\!\pm\!1.45$ & $52.43\!\pm\!1.68$ & $46.31\!\pm\!0.92$ & $42.94\!\pm\!0.46$ & $50.59\!\pm\!0.52$ & $51.03\!\pm\!1.01$ \\
PEAFusion-B~\cite{peafusion} & $56.44\!\pm\!1.72$ & $55.07\!\pm\!1.85$ & $40.45\!\pm\!1.44$ & $39.61\!\pm\!1.28$ & $45.40\!\pm\!3.52$ & $49.71\!\pm\!2.04$ \\
SpatialSeg (clean)~\cite{spatialseg} & $\underline{57.44}\!\pm\!0.36$ & $\underline{56.16}\!\pm\!0.08$ & $\underline{52.17}\!\pm\!0.98$ & $\underline{49.74}\!\pm\!0.16$ & $\underline{52.27}\!\pm\!0.60$ & $\underline{55.27}\!\pm\!0.25$ \\
ReSUL & $57.38\!\pm\!0.52$ & $\mathbf{56.82}\!\pm\!0.55$ & $\mathbf{53.96}\!\pm\!0.30$ & $\mathbf{51.24}\!\pm\!0.51$ & $\mathbf{53.43}\!\pm\!0.21$ & $\mathbf{55.93}\!\pm\!0.25$ \\
\midrule
\multicolumn{7}{l}{ImageNet-22k} \\
PEAFusion-B~\cite{peafusion} & $56.63\!\pm\!1.65$ & $54.67\!\pm\!1.08$ & $43.91\!\pm\!1.35$ & $37.56\!\pm\!2.45$ & $46.84\!\pm\!2.06$ & $49.22\!\pm\!0.75$ \\
PEAFusion-L~\cite{peafusion} & $\underline{58.19}\!\pm\!1.67$ & $56.50\!\pm\!1.43$ & $45.21\!\pm\!1.71$ & $35.16\!\pm\!2.33$ & $49.02\!\pm\!1.93$ & $50.48\!\pm\!1.53$ \\
SpatialSeg (clean)~\cite{spatialseg} & $57.56\!\pm\!1.55$ & $\underline{56.83}\!\pm\!1.89$ & $\underline{52.60}\!\pm\!3.18$ & $\underline{51.34}\!\pm\!1.77$ & $\underline{53.36}\!\pm\!1.61$ & $\underline{55.72}\!\pm\!1.42$ \\
ReSUL & $\mathbf{58.31}\!\pm\!0.51$ & $\mathbf{57.81}\!\pm\!0.28$ & $\mathbf{54.78}\!\pm\!0.99$ & $\mathbf{52.12}\!\pm\!0.64$ & $\mathbf{54.27}\!\pm\!0.54$ & $\mathbf{56.60}\!\pm\!0.57$ \\
\bottomrule
\end{tabular*}
\par\smallskip
\begin{tabular*}{\textwidth}{@{\extracolsep{\fill}}l*{5}{c}@{}}
\toprule
\multicolumn{6}{l}{\textit{Thermal perturbations}} \\
Method & Shift & Severe & Rotation & Local shift & Off \\
\midrule
\multicolumn{6}{l}{ImageNet-1k} \\
CMX-B2~\cite{cmx} & $41.63\!\pm\!0.55$ & $37.73\!\pm\!0.51$ & $43.18\!\pm\!0.41$ & $51.22\!\pm\!0.61$ & $43.53\!\pm\!2.14$ \\
CMNeXt-B4~\cite{deliver} & $\underline{47.71}\!\pm\!0.93$ & $45.16\!\pm\!2.48$ & $48.03\!\pm\!1.64$ & $\underline{54.26}\!\pm\!0.30$ & $45.59\!\pm\!6.80$ \\
AMDANet~\cite{amdanet2025} & $44.47\!\pm\!1.43$ & $42.07\!\pm\!1.32$ & $45.85\!\pm\!1.37$ & $52.12\!\pm\!0.95$ & $48.70\!\pm\!1.53$ \\
PEAFusion-B~\cite{peafusion} & $47.27\!\pm\!1.89$ & $\underline{47.38}\!\pm\!1.78$ & $\underline{48.18}\!\pm\!1.13$ & $52.97\!\pm\!1.90$ & $50.49\!\pm\!2.44$ \\
SpatialSeg (clean)~\cite{spatialseg} & $42.17\!\pm\!0.59$ & $37.81\!\pm\!0.56$ & $43.58\!\pm\!1.33$ & $50.93\!\pm\!0.52$ & $\underline{52.15}\!\pm\!0.38$ \\
ReSUL & $\mathbf{55.58}\!\pm\!0.25$ & $\mathbf{54.58}\!\pm\!0.42$ & $\mathbf{54.13}\!\pm\!0.64$ & $\mathbf{54.57}\!\pm\!0.40$ & $\mathbf{53.75}\!\pm\!0.83$ \\
\midrule
\multicolumn{6}{l}{ImageNet-22k} \\
PEAFusion-B~\cite{peafusion} & $47.51\!\pm\!1.07$ & $47.61\!\pm\!1.68$ & $48.12\!\pm\!1.94$ & $53.67\!\pm\!1.88$ & $49.99\!\pm\!2.09$ \\
PEAFusion-L~\cite{peafusion} & $\underline{50.43}\!\pm\!0.96$ & $\underline{50.00}\!\pm\!0.72$ & $\underline{51.02}\!\pm\!0.76$ & $\mathbf{55.48}\!\pm\!1.30$ & $52.35\!\pm\!1.10$ \\
SpatialSeg (clean)~\cite{spatialseg} & $43.42\!\pm\!1.04$ & $39.27\!\pm\!0.72$ & $45.28\!\pm\!1.68$ & $51.06\!\pm\!1.17$ & $\underline{53.51}\!\pm\!1.79$ \\
ReSUL & $\mathbf{56.44}\!\pm\!0.55$ & $\mathbf{55.63}\!\pm\!0.40$ & $\mathbf{55.22}\!\pm\!0.54$ & $\underline{55.45}\!\pm\!0.51$ & $\mathbf{54.39}\!\pm\!0.70$ \\
\bottomrule
\end{tabular*}
\end{table*}
ReSUL leads all ten perturbed conditions in the retrained
ImageNet-1k group (Table~\ref{tab:retrained}). Under severe thermal
displacement, it reaches 54.58\% mIoU, compared with 47.38\% for the
strongest competing method in this condition, PEAFusion-B, and 37.81\%
for the clean-trained reference. Clean mIoU remains close to the
reference, at 57.38\% versus 57.44\%. The improvement therefore extends
across the disturbance panel while retaining the reference's normal-input
performance to within 0.1 percentage point.

The ImageNet-22k comparison gives the same broad pattern: first place
in ten of eleven conditions, and second on localized thermal
displacement, where PEAFusion-L leads by 0.02 percentage point.
Severe-displacement mIoU reaches 55.63\%, versus
50.00\% for PEAFusion-L and 39.27\% for the reference. Clean mIoU rises
from the reference's 57.56\% to 58.31\%. Leadership across camera noise,
blur, and blackout as well as geometric faults shows that the retrained
bank supports more than displacement correction alone.

The three-run AMDANet comparison further shows how rankings change
with the input condition. Its clean mean is 57.08\%, close to ReSUL's
57.38\%, whereas severe displacement reduces it to 42.07\%. ReSUL
maintains 54.58\% in this condition and higher means across all eleven
conditions. The separation under disturbance is substantially larger
than the difference on clean images.

\subsection{Contributions of the Paths and Utility Learning}
\begin{figure*}[t]
\centering
\includegraphics[width=\textwidth]{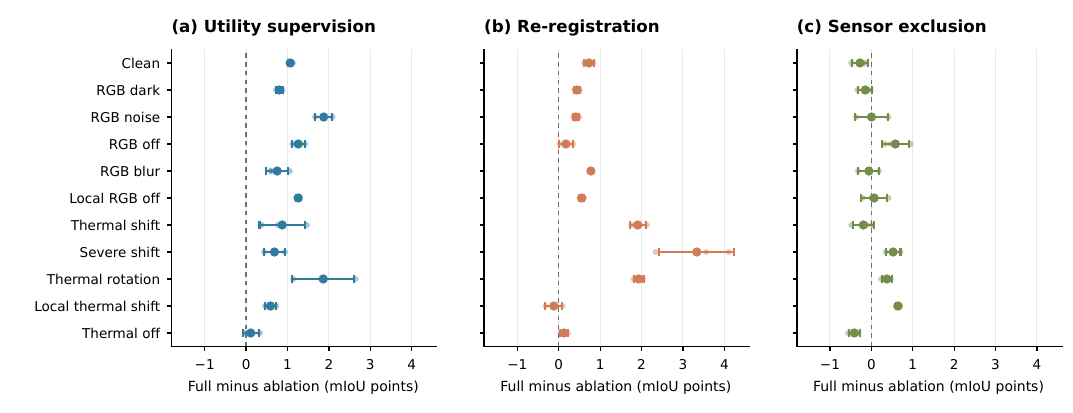}
\caption[Component contributions across disturbances]{Contribution of each component across all MFNet conditions in the retrained R1 bank. Positive differences favor the full model. Large points and bars show the mean and SD of three matched-run differences; faint points show individual runs.}
\label{fig:transitions}
\end{figure*}

\begin{table*}[t]
\centering
\caption{Component ablations (mIoU, \%). MFNet uses the fixed R1 bank; learned variants report mean $\pm$ SD over three router runs. Deterministic paths have no run SD.}
\label{tab:factors}
\footnotesize
\setlength{\tabcolsep}{2pt}
\begin{tabular*}{\textwidth}{@{\extracolsep{\fill}}l*{6}{c}@{}}
\toprule
\multicolumn{7}{l}{\textit{MFNet: fixed R1 bank}} \\
Method & Clean & RGB noise & RGB off & Severe shift & Rotation & Local shift \\
\midrule
Clean reference & $57.65$ & $51.61$ & $49.65$ & $38.13$ & $42.81$ & $51.23$ \\
Re-registration only & $57.08$ & $53.48$ & $48.87$ & $\underline{54.45}$ & $54.31$ & $53.54$ \\
Uniform allocation & $57.31$ & $\mathbf{54.51}$ & $50.36$ & $47.10$ & $52.07$ & $53.36$ \\
w/o re-registration & $56.97\!\pm\!0.10$ & $53.68\!\pm\!0.14$ & $\underline{50.60}\!\pm\!0.31$ & $51.47\!\pm\!0.80$ & $52.77\!\pm\!0.18$ & $\mathbf{55.11}\!\pm\!0.18$ \\
w/o sensor exclusion & $\mathbf{57.97}\!\pm\!0.02$ & $\underline{54.09}\!\pm\!0.19$ & $50.19\!\pm\!0.18$ & $54.28\!\pm\!0.10$ & $\underline{54.32}\!\pm\!0.26$ & $54.35\!\pm\!0.07$ \\
w/o utility supervision & $56.63\!\pm\!0.17$ & $52.22\!\pm\!0.02$ & $49.52\!\pm\!0.31$ & $54.12\!\pm\!0.29$ & $52.83\!\pm\!0.65$ & $54.41\!\pm\!0.13$ \\
ReSUL & $\underline{57.70}\!\pm\!0.18$ & $\underline{54.09}\!\pm\!0.21$ & $\mathbf{50.78}\!\pm\!0.25$ & $\mathbf{54.80}\!\pm\!0.13$ & $\mathbf{54.69}\!\pm\!0.16$ & $\underline{55.00}\!\pm\!0.04$ \\
\bottomrule
\end{tabular*}
\end{table*}
Utility supervision improves allocation even when the available experts
are held fixed. In the retrained R1 bank, it increases mean mIoU over
task-only routing in all eleven conditions (Table~\ref{tab:factors}
and Fig.~\ref{fig:transitions}). Clean mIoU rises from 56.63\% to
57.70\%, and RGB-noise mIoU from 52.22\% to 54.09\%. Because these
routers use the same predictions, the difference comes from how
allocation is learned, rather than from a stronger expert.

Re-registration supplies the largest path-specific gain under severe
displacement. Removing it lowers R1 mIoU from 54.80\% to 51.47\%.
The re-registered path alone reaches 54.45\% in this condition.
Allocation adds value across other faults: the full
bank improves on this path under noise, blackout, and localized
displacement. Repair and routing therefore have distinct roles,
recovering useful geometry and selecting among the available predictions.

Sensor exclusion provides a separate response to unusable inputs.
Removing its paths lowers R1 camera-blackout mIoU from 50.78\% to
50.19\%, while improving clean performance. Its contribution depends
on the disturbance, as the all-condition comparison below shows.
The paired differences in Fig.~\ref{fig:transitions} reveal different
patterns across the eleven R1 conditions. Utility supervision gives
positive mean differences throughout; re-registration has its largest
effect under severe displacement. Exclusion brings smaller gains under
camera blackout and localized thermal displacement, with decreases on
clean inputs and thermal blackout. Thus, a path's value depends on the
input, which is precisely the distinction the router is trained to learn.

\subsection{Predicting the Benefit of Sensor Exclusion}
\begin{table}[t]
\centering
\caption{Sensor-exclusion utility over the core conditions: balanced accuracy and recalls (\%), and raw-risk RMSE. Learned predictors use mean $\pm$ SD over three router runs.}
\label{tab:supervision}
\scriptsize
\setlength{\tabcolsep}{2pt}
\begin{tabular*}{\columnwidth}{@{\extracolsep{\fill}}lcccc@{}}
\toprule
Predictor & BAcc$\uparrow$ & $R_+\uparrow$ & $R_-\uparrow$ & RMSE$\downarrow$ \\
\midrule
\multicolumn{5}{@{}l}{\emph{MFNet, retrained R1: thermal exclusion}} \\
Entropy & $66.40$ & $38.53$ & $\mathbf{94.26}$ & -- \\
Task-only & $\underline{66.47}\!\pm\!0.73$ & $\mathbf{61.75}\!\pm\!6.88$ & $71.18\!\pm\!7.43$ & $\underline{0.1871}\!\pm\!0.0020$ \\
Utility & $\mathbf{67.51}\!\pm\!1.23$ & $\underline{52.89}\!\pm\!1.34$ & $\underline{82.14}\!\pm\!1.51$ & $\mathbf{0.1337}\!\pm\!0.0014$ \\
\midrule
\multicolumn{5}{@{}l}{\emph{MFNet, retrained R1: RGB exclusion}} \\
Entropy & $\mathbf{64.62}$ & $\underline{39.83}$ & $\mathbf{89.41}$ & -- \\
Task-only & $\underline{64.37}\!\pm\!3.73$ & $\mathbf{47.82}\!\pm\!14.95$ & $80.92\!\pm\!7.56$ & $\underline{0.1707}\!\pm\!0.0132$ \\
Utility & $59.23\!\pm\!3.09$ & $30.24\!\pm\!4.42$ & $\underline{88.23}\!\pm\!6.96$ & $\mathbf{0.1138}\!\pm\!0.0033$ \\
\bottomrule
\end{tabular*}
\end{table}
Utility supervision makes the predicted benefit of exclusion more
accurate in magnitude (Table~\ref{tab:supervision}). In the fixed R1
bank, it lowers raw-risk RMSE from 0.1871 to 0.1337 for thermal
exclusion and from 0.1707 to 0.1138 for RGB exclusion, reductions of
28.5\% and 33.3\%. The eleven-condition aggregates in Appendix~\ref{app:utility}
show the same pattern. These comparisons connect the learned scores
to the risk differences used as supervision.

Benefit magnitude and sign measure different aspects of this prediction.
Thermal-exclusion balanced accuracy increases from 66.47\% to 67.51\%.
For RGB, retention recall rises from 80.92\% to 88.23\%, while
exclusion recall falls from 47.82\% to 30.24\%. Thus, lower regression
error does not imply better binary decisions at every threshold.
ReSUL uses the continuous utilities to form a spatial mixture, and
the same-bank segmentation gains in Table~\ref{tab:factors} establish
the contribution of this supervision to the final prediction.

\subsection{Regional Allocation in Individual Scenes}
\begin{figure*}[tp]
\centering
\includegraphics[width=\textwidth]{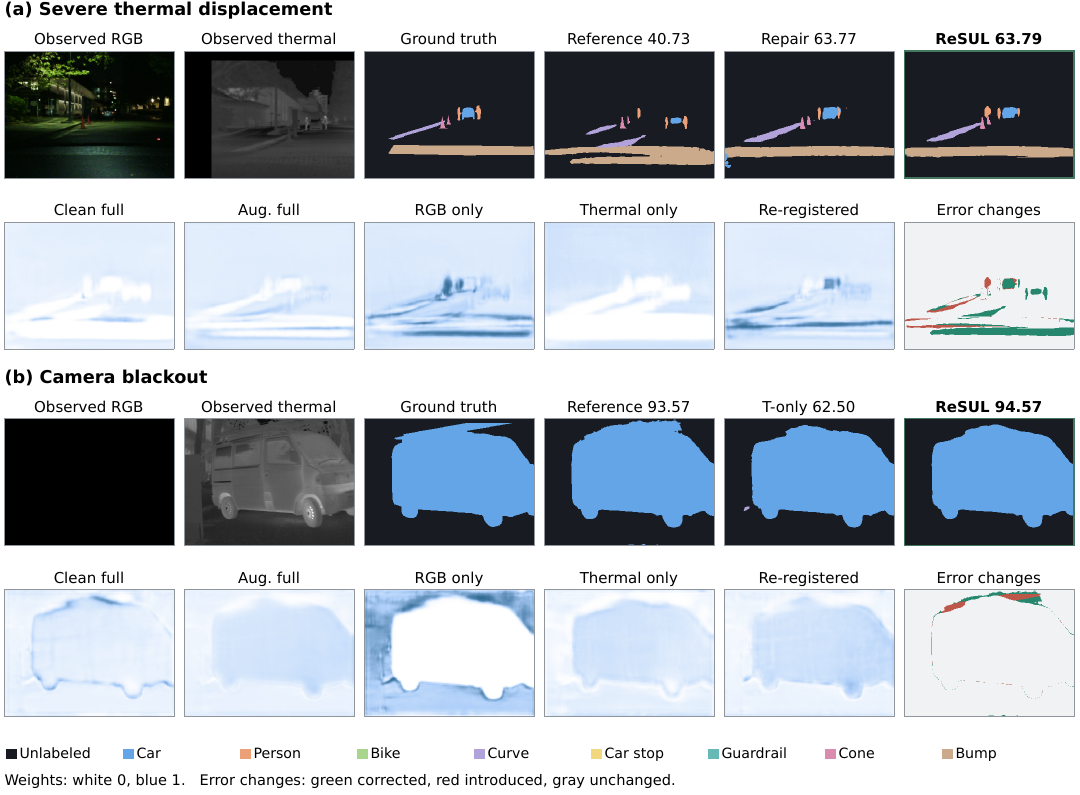}
\caption[Predictions and regional utility fields]{Regional allocation under (a) severe thermal displacement and (b) camera blackout on MFNet. Each block shows the observations, ground truth, reference, a condition-relevant alternative, ReSUL, and all five path weights. Predictions use the project-trained R1 bank and router initialization 7. Numbers denote per-image mIoU over nonzero-union classes; weight maps share $[0,1]$. Error changes distinguish corrected pixels (green) from newly introduced errors (red) relative to the reference.}
\label{fig:qualitative}
\end{figure*}
The severe-displacement example in Fig.~\ref{fig:qualitative}(a)
shows that useful thermal structure survives misalignment. The
re-registered path recovers the horizontal speed-bump region and
raises per-image mIoU from the reference's 40.73\% to 63.77\%.
ReSUL reaches 63.79\%, with spatially varying contributions from
the repaired and sensor-exclusion paths. The error-change map locates
the corrections rather than reducing them to a single image score.

Camera blackout requires a different response
(Fig.~\ref{fig:qualitative}(b)). ReSUL reduces the excess vehicle
region in the reference prediction and raises per-image mIoU from
93.57\% to 94.57\%. The shared weight scale reveals how several
paths contribute within the scene. These examples complement the
dataset-level ablations by locating where a changed prediction helps
and where it introduces an error.
\section{Discussion}
\label{sec:discussion}
The results distinguish the availability of a useful prediction from the
ability to select it. Re-registration recovers scene structure that
would otherwise be discarded, and exclusion supplies predictions that
do not depend on a disturbed input. Utility learning then determines
where these alternatives should contribute. This separation explains
why an effective repair path can drive displacement robustness while
same-bank utility supervision still improves allocation across faults.
Sensor usefulness is consequently tied to a prediction choice.
Displaced thermal structure can regain value after repair, while
an unavailable camera calls for a path that retains thermal evidence.
The fixed-bank comparisons show that these alternatives are most useful
when allocation is learned from their regional task benefit. The
calibration results also distinguish the accuracy of a continuous
utility estimate from the accuracy of a binary exclusion decision:
ReSUL learns a mixture, not a universal sensor-rejection threshold.

The present routers are fitted per sensor configuration using synthetic
disturbances. Recorded sensor faults and calibration drift over time
provide natural next tests of the same regional selection principle.
\section{Conclusion}
ReSUL learns how to use disturbed sensor observations by
comparing the regional prediction benefit of retaining, excluding, or
repairing them. Sensor-conditioned prediction paths provide these alternatives, and
signed risk supervision connects their allocation to segmentation quality.
MFNet experiments with project-trained models show leading mean mIoU
in ten of eleven conditions in each initialization group. Fixed-bank
ablations identify the separate roles of geometric repair, sensor
exclusion, and utility supervision; regional risk diagnostics connect
the learned scores to the benefit of these prediction choices.
The central finding is that useful information can survive sensor
interference. Robust fusion benefits from recovering that information
and learning where its use improves the prediction.

\appendices
\setcounter{table}{0}
\renewcommand{\thetable}{S\arabic{table}}
\renewcommand{\theHtable}{appendix.\arabic{table}}
\section{Training and Aggregation}
\label{app:training}
The MFNet expert networks use the SpatialSeg architecture with
ImageNet-1k or ImageNet-22k initialization. Local training executes
27,500 optimizer steps with a global batch of six; AutoDL training
executes 16,500 steps with a global batch of ten. Both expose an expert
to 165,000 sampled training examples. Their polynomial learning-rate
horizons are 116,667 and 70,000 steps, respectively. The horizon defines
the decay schedule rather than the number of executed optimizer steps.
Standard augmentations are used by the clean reference. The alternative
expert additionally samples geometric and appearance faults; external
methods retain their method-specific training recipes. Checkpoint
selection uses the official validation split.

The retrained 1k and 22k results summarize three and five runs,
respectively. Within each run, the condition score averages router
initializations 7, 22, and 42. The reported SD is the sample SD of these
run means. Fixed-R1 ablations instead summarize the three
router initializations of their stated bank. Deterministic paths and
single-checkpoint baselines do not receive an artificial run SD. The
source records retain checkpoint identifiers, run membership, condition
scores, and the exact aggregation used for each table cell.

In the retrained bank, the full-sensor clean expert is the reference;
the fault-augmented expert contributes its full-sensor, camera-only,
thermal-only, and re-registered-thermal predictions. A same-parent
exclusion comparison pairs the latter network's full and restricted
paths.

AMDANet uses its official MiT-B4 implementation and ImageNet-1k
initialization. Each of three runs trains for 350 epochs (45,850
optimizer updates), with batch size six, AdamW, initial learning rate
$6.3\times10^{-5}$, weight decay 0.01, 50 warm-up epochs, and polynomial
decay with exponent 0.9. The best validation checkpoint is evaluated
in FP32 without test-time augmentation on the same native-resolution
disturbances as ReSUL. The CMX-B2, CMNeXt-B4, and PEAFusion-B 1k
comparisons use three runs; PEAFusion-B/L 22k use five runs. All
reported dispersions are sample SDs of the recorded run scores.
CMX-B2 and CMNeXt-B4 train for 500 epochs with global batch eight.
PEAFusion-B/L train for 35,000 optimizer updates with global batches
14/16, respectively; the B-1k, B-22k, and L-22k configurations use
attention windows 16, 12, and 12.

\section{Complete Component Ablations}
Table~\ref{tab:allablations} reports all eleven MFNet conditions for
the component comparisons summarized in the main paper. The retrained
analysis uses the same R1 bank throughout. Figure~\ref{fig:transitions}
uses paired router-initialization differences from these records,
with sample SD rather than a confidence interval.
\begin{table*}[t]
\centering
\caption{Complete fixed-R1 MFNet ablations (mIoU, \%). Learned variants use three router runs.}
\label{tab:allablations}
\footnotesize
\setlength{\tabcolsep}{2pt}
\begin{tabular*}{\textwidth}{@{\extracolsep{\fill}}l*{6}{c}@{}}
\toprule
\multicolumn{7}{l}{\textit{MFNet retrained R1: original inputs and camera perturbations}} \\
Method & Clean & Dark & Noise & Off & Blur & Local off \\
\midrule
Clean reference & $57.65$ & $56.20$ & $51.61$ & $49.65$ & $51.93$ & $55.42$ \\
Re-registration only & $57.08$ & $57.06$ & $53.48$ & $48.87$ & $53.69$ & $55.10$ \\
Uniform allocation & $57.31$ & $57.07$ & $\mathbf{54.51}$ & $50.36$ & $\mathbf{54.00}$ & $55.48$ \\
w/o re-registration & $56.97\!\pm\!0.10$ & $56.98\!\pm\!0.06$ & $53.68\!\pm\!0.14$ & $\underline{50.60}\!\pm\!0.31$ & $52.87\!\pm\!0.14$ & $55.49\!\pm\!0.18$ \\
w/o sensor exclusion & $\mathbf{57.97}\!\pm\!0.02$ & $\mathbf{57.56}\!\pm\!0.06$ & $\underline{54.09}\!\pm\!0.19$ & $50.19\!\pm\!0.18$ & $\underline{53.70}\!\pm\!0.12$ & $\underline{55.98}\!\pm\!0.18$ \\
w/o utility supervision & $56.63\!\pm\!0.17$ & $56.61\!\pm\!0.17$ & $52.22\!\pm\!0.02$ & $49.52\!\pm\!0.31$ & $52.89\!\pm\!0.32$ & $54.79\!\pm\!0.20$ \\
ReSUL & $\underline{57.70}\!\pm\!0.18$ & $\underline{57.41}\!\pm\!0.11$ & $\underline{54.09}\!\pm\!0.21$ & $\mathbf{50.78}\!\pm\!0.25$ & $53.64\!\pm\!0.14$ & $\mathbf{56.04}\!\pm\!0.16$ \\
\bottomrule
\end{tabular*}
\par\smallskip
\begin{tabular*}{\textwidth}{@{\extracolsep{\fill}}l*{5}{c}@{}}
\toprule
\multicolumn{6}{l}{\textit{MFNet retrained R1: thermal perturbations}} \\
Method & Shift & Severe & Rotation & Local shift & Off \\
\midrule
Clean reference & $42.51$ & $38.13$ & $42.81$ & $51.23$ & $52.37$ \\
Re-registration only & $\mathbf{55.82}$ & $\underline{54.45}$ & $54.31$ & $53.54$ & $53.49$ \\
Uniform allocation & $52.20$ & $47.10$ & $52.07$ & $53.36$ & $53.76$ \\
w/o re-registration & $53.46\!\pm\!0.19$ & $51.47\!\pm\!0.80$ & $52.77\!\pm\!0.18$ & $\mathbf{55.11}\!\pm\!0.18$ & $54.20\!\pm\!0.09$ \\
w/o sensor exclusion & $\underline{55.56}\!\pm\!0.17$ & $54.28\!\pm\!0.10$ & $\underline{54.32}\!\pm\!0.26$ & $54.35\!\pm\!0.07$ & $\mathbf{54.74}\!\pm\!0.05$ \\
w/o utility supervision & $54.50\!\pm\!0.48$ & $54.12\!\pm\!0.29$ & $52.83\!\pm\!0.65$ & $54.41\!\pm\!0.13$ & $54.21\!\pm\!0.12$ \\
ReSUL & $55.37\!\pm\!0.09$ & $\mathbf{54.80}\!\pm\!0.13$ & $\mathbf{54.69}\!\pm\!0.16$ & $\underline{55.00}\!\pm\!0.04$ & $\underline{54.32}\!\pm\!0.12$ \\
\bottomrule
\end{tabular*}
\end{table*}

\section{Five-Path Utility Diagnostics}
\label{app:utility}
The fixed MFNet R1 bank is evaluated on all
393 test images under eleven conditions. Each learned predictor uses
router initializations 7, 22, and 42; the tables report their mean and
sample SD after pooling region counts and errors within each run.
The entropy predictor is deterministic. The same-parent contrasts,
raw Brier-risk units, and $|H|\geq0.0025$ threshold match the main paper.
Table~\ref{tab:mfnet-utility-all11} extends the seven-condition main-table
aggregate with camera blur, auxiliary rotation, localized auxiliary
displacement, and localized camera blackout. The additional conditions
are held out from router fitting; rotation and localized displacement
are included in AlignNet and fault-augmented expert training.
\begin{table}[t]
\centering
\caption{MFNet five-path utility over all eleven conditions. Metrics follow the main-paper calibration table.}
\label{tab:mfnet-utility-all11}
\scriptsize
\setlength{\tabcolsep}{2pt}
\begin{tabular*}{\columnwidth}{@{\extracolsep{\fill}}lcccc@{}}
\toprule
Predictor & BAcc$\uparrow$ & $R_+\uparrow$ & $R_-\uparrow$ & RMSE$\downarrow$ \\
\midrule
\multicolumn{5}{@{}l}{\emph{MFNet, retrained R1: thermal exclusion}} \\
Entropy & $\underline{65.93}$ & $39.45$ & $\mathbf{92.41}$ & -- \\
Task-only & $63.73\!\pm\!0.64$ & $\mathbf{61.08}\!\pm\!7.23$ & $66.39\!\pm\!7.42$ & $\underline{0.1970}\!\pm\!0.0037$ \\
Utility & $\mathbf{65.94}\!\pm\!1.18$ & $\underline{52.96}\!\pm\!1.48$ & $\underline{78.92}\!\pm\!1.54$ & $\mathbf{0.1421}\!\pm\!0.0011$ \\
\midrule
\multicolumn{5}{@{}l}{\emph{MFNet, retrained R1: RGB exclusion}} \\
Entropy & $\mathbf{64.68}$ & $\underline{43.76}$ & $\underline{85.61}$ & -- \\
Task-only & $\underline{63.35}\!\pm\!3.06$ & $\mathbf{49.59}\!\pm\!14.55$ & $77.11\!\pm\!8.47$ & $\underline{0.1758}\!\pm\!0.0143$ \\
Utility & $59.63\!\pm\!2.72$ & $32.84\!\pm\!4.25$ & $\mathbf{86.42}\!\pm\!5.69$ & $\mathbf{0.1180}\!\pm\!0.0023$ \\
\bottomrule
\end{tabular*}
\end{table}

\section{Class-Level Segmentation}
Table~\ref{tab:cleanclasses} reports the complete clean-image class
metrics for AMDANet and ReSUL, including class zero. Values use the
same run aggregation as the main comparison. Condition-wise class
reports and confusion matrices are retained with the experiment records.
\begin{table*}[t]
\centering
\caption{Clean MFNet per-class results (\%, mean $\pm$ SD).}
\label{tab:cleanclasses}
\footnotesize
\setlength{\tabcolsep}{2pt}
\begin{tabular*}{\textwidth}{@{\extracolsep{\fill}}llcccc@{}}
\toprule
Method & Class & IoU & Precision & Recall & F1 \\
\midrule
AMDANet (1k, 3 runs) & Unlabeled & $98.09\pm0.10$ & $99.03\pm0.04$ & $99.04\pm0.13$ & $99.04\pm0.05$ \\
 & Car & $88.91\pm0.49$ & $94.08\pm0.43$ & $94.18\pm0.21$ & $94.13\pm0.27$ \\
 & Person & $73.93\pm0.60$ & $86.13\pm1.01$ & $83.93\pm1.34$ & $85.01\pm0.40$ \\
 & Bike & $62.87\pm1.83$ & $77.48\pm3.63$ & $77.06\pm3.05$ & $77.19\pm1.38$ \\
 & Curve & $47.41\pm0.47$ & $59.04\pm1.24$ & $70.69\pm1.64$ & $64.32\pm0.43$ \\
 & Car stop & $28.06\pm6.64$ & $59.25\pm15.00$ & $34.53\pm5.05$ & $43.54\pm8.15$ \\
 & Guardrail & $8.82\pm8.06$ & $9.67\pm8.26$ & $40.43\pm33.19$ & $15.56\pm13.24$ \\
 & Cone & $53.03\pm1.64$ & $83.68\pm2.74$ & $59.18\pm2.23$ & $69.29\pm1.41$ \\
 & Bump & $52.64\pm1.71$ & $67.91\pm3.29$ & $70.38\pm5.22$ & $68.97\pm1.48$ \\
\midrule
ReSUL (1k, 3 runs) & Unlabeled & $98.23\pm0.01$ & $99.28\pm0.01$ & $98.94\pm0.01$ & $99.11\pm0.00$ \\
 & Car & $90.66\pm0.13$ & $94.77\pm0.23$ & $95.43\pm0.24$ & $95.10\pm0.07$ \\
 & Person & $74.35\pm0.42$ & $83.12\pm1.20$ & $87.60\pm1.05$ & $85.29\pm0.27$ \\
 & Bike & $66.07\pm0.37$ & $76.74\pm1.29$ & $82.65\pm1.09$ & $79.57\pm0.27$ \\
 & Curve & $46.41\pm0.67$ & $57.78\pm2.63$ & $70.40\pm2.30$ & $63.40\pm0.62$ \\
 & Car stop & $25.12\pm1.82$ & $60.14\pm6.10$ & $30.16\pm1.49$ & $40.13\pm2.32$ \\
 & Guardrail & $5.36\pm0.76$ & $5.49\pm0.73$ & $67.73\pm11.24$ & $10.16\pm1.37$ \\
 & Cone & $56.54\pm0.78$ & $82.46\pm2.66$ & $64.32\pm1.26$ & $72.23\pm0.63$ \\
 & Bump & $53.65\pm1.67$ & $63.38\pm3.19$ & $78.13\pm6.09$ & $69.82\pm1.42$ \\
\midrule
ReSUL (22k, 5 runs) & Unlabeled & $98.26\pm0.05$ & $99.33\pm0.03$ & $98.92\pm0.08$ & $99.12\pm0.03$ \\
 & Car & $90.75\pm0.15$ & $94.53\pm0.52$ & $95.79\pm0.37$ & $95.15\pm0.08$ \\
 & Person & $74.15\pm0.34$ & $83.30\pm1.45$ & $87.13\pm1.42$ & $85.15\pm0.23$ \\
 & Bike & $65.73\pm0.34$ & $77.04\pm1.84$ & $81.81\pm1.66$ & $79.32\pm0.25$ \\
 & Curve & $48.03\pm1.19$ & $56.64\pm2.21$ & $76.06\pm1.33$ & $64.88\pm1.09$ \\
 & Car stop & $27.02\pm1.96$ & $60.16\pm1.57$ & $32.92\pm2.64$ & $42.50\pm2.36$ \\
 & Guardrail & $6.17\pm0.53$ & $6.33\pm0.54$ & $71.24\pm7.71$ & $11.62\pm0.95$ \\
 & Cone & $56.04\pm1.31$ & $81.95\pm2.54$ & $64.05\pm2.86$ & $71.82\pm1.08$ \\
 & Bump & $58.61\pm3.03$ & $67.10\pm0.86$ & $82.26\pm5.00$ & $73.86\pm2.46$ \\
\bottomrule
\end{tabular*}
\end{table*}

\section{Qualitative Examples}
MFNet examples use test images 01247N under severe thermal displacement
and 01562D under camera blackout, with the fixed R1 five-path router
initialization 7. Per-image mIoU in the visual panels averages the
classes with nonzero prediction--target union, including class zero.
Dataset-level tables instead use the full aggregated nine-class
confusion matrix. Expert-weight maps use one common zero-to-one scale.
\end{document}